\documentclass[10pt,twocolumn,letterpaper]{article}

\usepackage{cvpr}              

\usepackage[accsupp]{axessibility}

\usepackage{graphicx}
\usepackage{amsmath}
\usepackage{amssymb}
\usepackage{hhline, booktabs}
\usepackage{multirow}
\usepackage{algorithm}
\usepackage{algpseudocode}
\usepackage{comment}
\usepackage{xcolor,colortbl}
\definecolor{lavender}{rgb}{0.95,0.95,0.95}

\usepackage[pagebackref,breaklinks,colorlinks]{hyperref}

\usepackage[capitalize]{cleveref}
\crefname{section}{Sec.}{Secs.}
\Crefname{section}{Section}{Sections}
\Crefname{table}{Table}{Tables}
\crefname{table}{Tab.}{Tabs.}

\def\cvprPaperID{} 
\def\confName{CVPR}
\def\confYear{2023}

\begin{document}

\title{Coherent Concept-based Explanations in Medical Image and Its Application to Skin Lesion Diagnosis}

\author{Cristiano Patrício$^{1,3}$, João C. Neves$^{1}$, Luis F. Teixeira$^{2,3}$\\ \\
$^1$\textit{Universidade da Beira Interior and NOVA LINCS}, \\$^2$\textit{Faculdade de Engenharia da Universidade do Porto, $^3$INESC TEC}\\
}
\maketitle

\begin{abstract}

Early detection of melanoma is crucial for preventing severe complications and increasing the chances of successful treatment. Existing deep learning approaches for melanoma skin lesion diagnosis are deemed black-box models, as they omit the rationale behind the model prediction, compromising the trustworthiness and acceptability of these diagnostic methods. Attempts to provide concept-based explanations are based on post-hoc approaches, which depend on an additional model to derive interpretations. In this paper, we propose an inherently interpretable framework to improve the interpretability of concept-based models by incorporating a hard attention mechanism and a coherence loss term to assure the visual coherence of concept activations by the concept encoder, without requiring the supervision of additional annotations. The proposed framework explains its decision in terms of human-interpretable concepts and their respective contribution to the final prediction, as well as a visual interpretation of the locations where the concept is present in the image. Experiments on skin image datasets demonstrate that our method outperforms existing black-box and concept-based models for skin lesion classification.





\end{abstract}

\section{Introduction}
\label{sec:intro}

The research on automated medical image analysis is nowadays dominated by the use of deep learning models, which supported a remarkable increase in the accuracy of several medical image problems, assuring in some cases results that surpassed human performance~\cite{Majkowska_Radiology2020,Xu_Engineering2021}. In spite of these advances, these models are hardly ever adopted in clinical practice, and the explanation to this apparent paradox is straightforward, doctors will never trust the decision of an algorithm without understanding its decision process~\cite{lipton2017doctor}.
\begin{figure}[!htbp]
    \centering
    \includegraphics[width=0.4\textwidth]{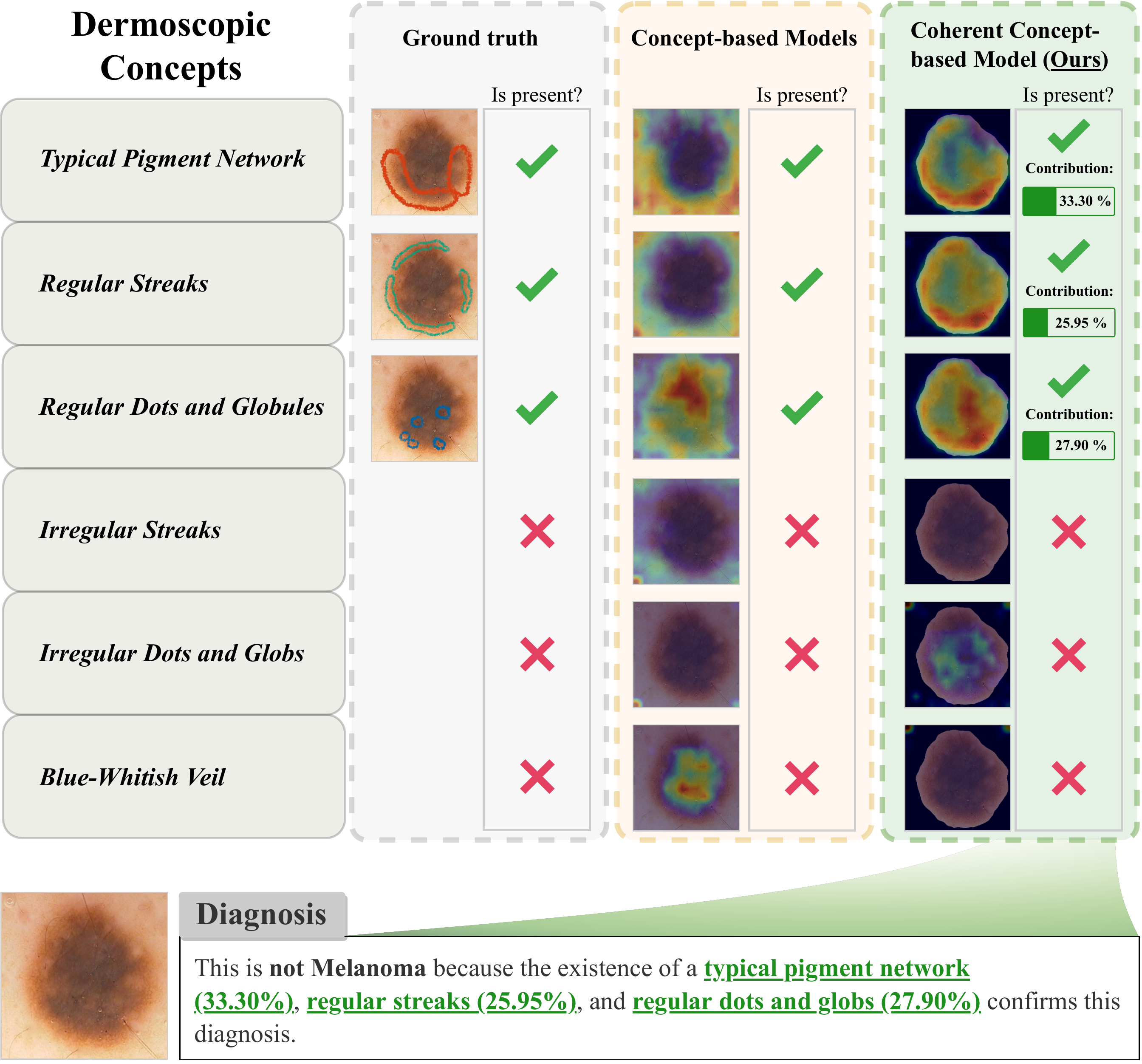}
    \caption{\textbf{Motivation of our proposal}. Our approach improves concept-based approaches for skin lesion diagnosis by enforcing the concept activations to be coherent with the locations that experts rely on for concept identification (compare the concept activations between Concept-based Models and \textit{Ours}).}
    \label{fig:intro}
\end{figure}
To address this problem, several strategies were proposed to understand the decision process of deep learning models through the use of saliency maps highlighting the contribution of each region/pixel in the prediction of the model. However, recent studies~\cite{adebayo2018sanity, Rudin_NMI2019} have demonstrated these strategies to be unreliable, subjective and meaningless for incorrect predictions. Moreover, Rudin~\cite{Rudin_NMI2019} has even urged researchers to stop using these methods for explaining the decisions of black-box models, arguing that inherently interpretable models can perform on par with black-box models. Based on this premise, researchers have been shifting their efforts to creating inherently interpretable models through the modification of the internal architecture or learning process of convolutional neural networks. Among the different inherently interpretable strategies, concept-based approaches are growing in popularity in medical imaging analysis~\cite{Fang_MM2020,Lucieri_IJCNN2020,Graziani_2018,chen_2020_NMI}, since they allow to decompose the decision process into a set of human-interpretable concepts. These approaches enforce specific filters of the network to be activated when a concept is present in an image and at the same time use them for class prediction, allowing to determine, during inference, the concepts that the network identified in the original image, as well as their contributions. Ideally, these approaches are expected to learn the locations to which each concept is related to in the original image and activate those corresponding regions in the feature maps of the concept layer. However, as depicted in Figure~\ref{fig:intro}, regions highlighted by each concept are often incongruent with the locations where the concept is present. We argue that this effect is caused by the lack of a mechanism that guides concept activations to focus on relevant regions on the image.

In this paper, we introduce a strategy that is capable of enforcing the visual coherence of concept activations without requiring supervision of the concept location masks, avoiding the need to obtain these annotations from experts. Specifically, a coherence loss term and a preprocessing strategy are proposed to guide the activations of concept filters towards the locations where to which the concept is visually related to. A  general concept-based CNN ~\cite{wickramanayake2021comprehensible} is adopted as the base architecture, comprising a concept layer with $k$ filters representing each of the human-interpretable concepts. During the learning phase, the proposed learning strategy enforces the concept layer filters to activate when the concept is present in the image, as well as to concentrate the activations on the locations that the concept is related to. For defining the regions of interest without requiring additional annotations from experts, we rely on a standard image segmentation approach trained on external data and we show that this strategy performs on par with the use of manually defined regions.
Extensive experiments on existing skin lesions datasets demonstrate that the proposed approach can not only improve the interpretability of the diagnosis, but also the classification accuracy of the model, as evidenced by the increase on the accuracy of concept identification and skin lesion classification. Our contributions can be summarized as follows: 1) we introduce a strategy to improve the interpretability of concept-based models for skin lesion diagnosis through the enforcement of concepts activations consistent with the image regions to which the concepts are related to; 2) we show that the proposed strategy is capable of not only improving individual concept interpretation but also improves the overall classification accuracy of skin lesion classification; 3) our strategy was designed to avoid the need of additional annotations by exploiting a standard image segmentation trained on external data. 


\section{Related Work}
\label{sec:related_work}

\paragraph{Concept-Based Models}

The rationale behind concept-based learning approaches is using human-specified concepts as an intermediate step to derive the final predictions. This idea was initially explored in~\cite{Kumar_CVPR_2009} and~\cite{lampert2009learning} for few-shot learning approaches. More recently, deep neural networks with concept bottlenecks~\cite{koh_2020_ICML, Kazhdan_2020_AIMLAI, chen_2020_NMI, wickramanayake2021comprehensible} have re-emerged for solving a panoply of tasks. These Concept Bottleneck Models~\cite{koh_2020_ICML} (CBM) relied on an encoder-decoder paradigm, where the encoder is responsible for predicting the concepts given the input image, and the decoder leverages the predicted concepts to infer the final predictions. Wickramanayake \etal~\cite{wickramanayake2021comprehensible} extended the baseline idea of CBM by leveraging semantic information of the concepts to preserve the distance between the visual features of the concept layer and the corresponding concept semantic representation in a joint embedding space. More recently, Sarkar \etal~\cite{sarkar_2022_CVPR} proposed an ante-hoc explainable model in which the output of the concept encoder is passed to a decoder that reconstructs the input image, encouraging the model to capture the semantic features of the input image. These works targeted classification tasks in general images, and there was no attempt to develop a concept-based, inherently interpretable model for automatic skin lesion diagnosis. To this end, we introduce a strategy to improve the interpretability of existing concept-based models by inserting a coherence loss that encourages the concept activations to occur within the image regions in which the concepts are present.

\paragraph{Interpretable Skin Lesion Classification}

The typical procedure physicians adopt for skin lesion diagnosis is the observation through naked-eye or dermoscopic imaging using manual algorithms, such as the ABCD-rule~\cite{Nachbar1994JAAD} or 7-point checklist~\cite{argenziano1998epiluminescence}, which requires a certain level of expertise and experience. Automatic approaches for classifying skin lesions include the traditional deep learning methods~\cite{esteva2017dermatologist, liu2020deep}. However, these methods are considered black-box models. Therefore, explainable approaches for dermoscopic skin lesion diagnosis have emerged and they typically rely on applying post-hoc visual methods such as saliency maps~\cite{xiang2019towards} or attention mechanisms~\cite{barata2021explainable}. However, these approaches explain the model decision in a post-hoc manner and may often produce ambiguous explanations. To the best of our knowledge, Lucieri \etal~\cite{lucieri2022exaid} were the first to propose a framework providing multimodal concept-based explanations for melanoma classification. Their framework relied on post-hoc methods~\cite{kim2018interpretability} for concept identification and for estimating the influence of a specific concept on the decision to produce textual explanations. Additionally, the authors used the Concept Localization Maps~\cite{lucieri2020explaining} to show the location of the concepts in the image. In contrast, our method does not requires additional effort to obtain interpretation after training. Herein, we aim to build a framework that is end-to-end trainable and interpretable by design, where predictions are solely based on the concept information from the concept encoder.

\section{Method}
\label{sec:method}

Inspired by~\cite{wickramanayake2021comprehensible}, we designed a model that is capable of explaining its decision process using human-interpretable concepts. Interpretability is achieved by estimating the concept contribution to the final decision, as well as, the locations in the image where the concept is present. For this, as depicted in Figure~\ref{fig:method}, a concept encoder layer is added on top of a standard feature extractor, which is used to determine the presence of a specific concept in the image, allowing to disentangle the decision process carried out by the fully connected layers of the model. Additionally, the feature maps produced by the concept layer are enforced to highlight the regions in the image to which the concept is associated using simultaneously a mapping consistency loss and a coherence loss. The following sections describe the different parts of the proposed framework.

\subsection{Framework Architecture}

\begin{figure*}[h!]
    \centering
    \includegraphics[width=\textwidth]{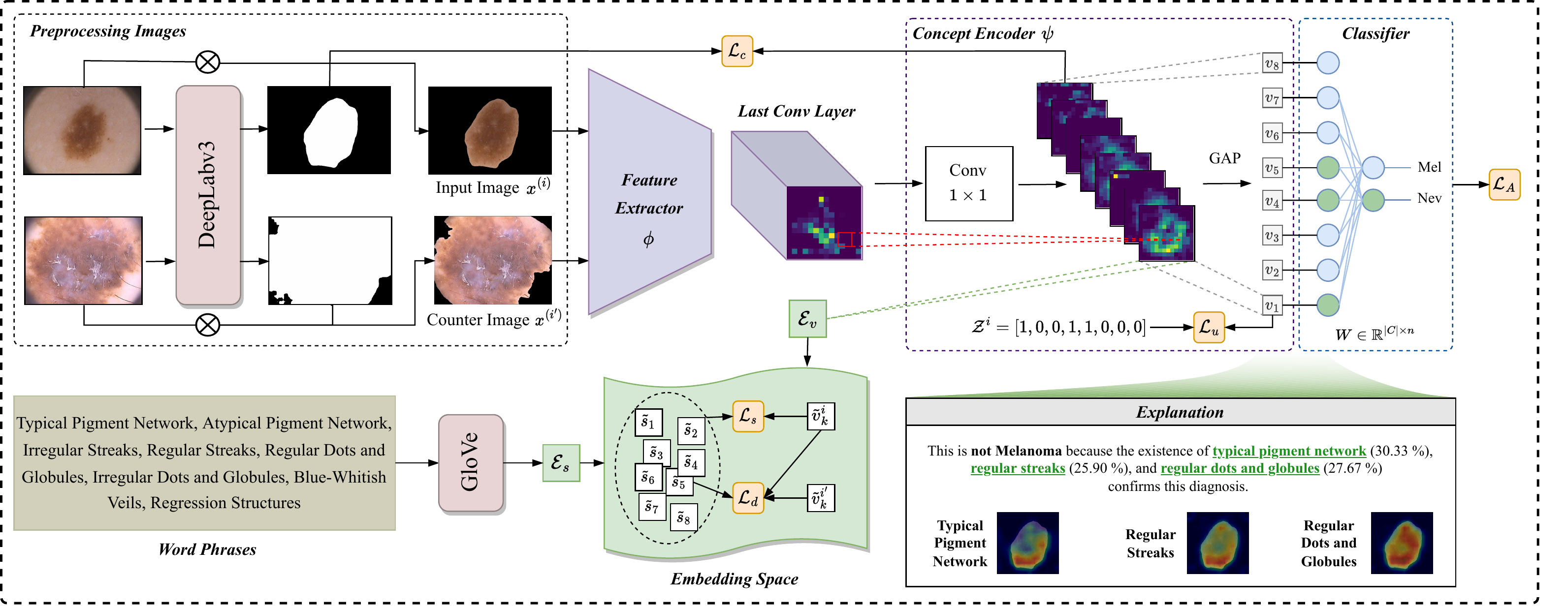}
    \caption{\textbf{Overview of the proposed framework.} Our approach is divided into two major phases: 1) The pre-processing stage uses a semantic segmentation model for obtaining the lesion segmentation mask, which is subsequently used to mask the input image. 2) In the learning phase, a concept encoder is used to disentangle the feature maps of a CNN into human-interpretable concepts. Each concept is encoded by a filter whose activations are enforced to be activated in the presence of the concept in the original image (uniqueness loss), share similarities with concepts close in the textual embedding space (mapping loss), and produce visual activations consistent with the locations of the concept in the image (coherence loss). During inference, the trained model not only predicts the class label but also estimates the dermoscopic concepts and their respective contribution (in \%) to the final decision, in addition to the produced feature maps by the concept encoder, which allows visualizing the location of each concept in the image.}
    \label{fig:method}
\end{figure*}

Given a standard Deep Neural Network (DNN) classifier $f:  \mathbf{\mathcal{X}} \to  \mathcal{Y}$, $ \mathcal{X} \in \mathbb{R}^n, \mathcal{Y} \in \{0, 1\}$, we consider the feature extractor $\phi: \mathcal{X} \to \mathcal{F}_x^l$, which takes an image $x^{i}$ as input and outputs a set of feature maps $\mathcal{F}_x^l$ of the layer $l$, typically the last convolutional layer. A concept encoder $\psi$ is then incorporated on top of the feature extractor $\phi$ to disentangle the learned feature representations into a set of intermediate activations related to human-interpretable concepts. The concept encoder $\phi$ takes as input the feature maps $\mathcal{F}_x^l$ and maps it to the concepts $\textbf{c} = \{c^1, ..., c^k\}$, resulting in $k$ feature maps $\mathcal{A}_k^{i}$, where $k$ is the number of concepts. Moreover, each concept $c^k$ has a semantic description, e.g. ``Typical Pigment Network'', which is encoded by the GloVe~\cite{pennington2014glove} model to have a vector representation $\textbf{s} \in \mathcal{\mathbb{R}}^d$, which we denote as word phrases. This semantic information will contribute to each filter in the concept encoder to learn a unique concept. Finally, we rely on a classifier module, composed of one FC layer to map the concept activations to the final class label $\hat{y}$. The general scheme of the proposed framework is depicted in Figure \ref{fig:method}.

\subsection{Concept Encoder}

The concept encoder $\psi$ is composed of a convolutional layer with a kernel size of $1 \times 1$, which takes as input the result of $\phi(x^{i})$ and outputs $k$ feature maps $\mathcal{A}^{i}$, followed by a Rectified Linear Unit (ReLU) layer to ensure only positive contributions, and a dropout layer during training. The core idea of the concept encoder is to guide the convolutional filters to learn human-interpretable concepts. Let the visual feature $v_k^i = \frac{1}{p.q} \sum_a^p \sum_b^q \mathcal{A}_{ab}^k$ be the result of a Global Average Pooling (GAP) operation over the feature map $\mathcal{A}_k^{i}$. 
There should be a one-to-one association between the visual feature $v_k^i$ and the word phrase $s_k$. This means that each filter is responsible for learning a unique concept. To achieve these properties, a set of regularization terms are defined and discussed in section \ref{sec:loss}.

\subsection{Objective Function}
\label{sec:loss}

\paragraph{Concept Uniqueness Loss}

Explaining the model decision based on human-interpretable concepts is achieved by mapping the high-level features of the last convolutional layer to a set of scalar values denoting the presence/absence of a concept. To learn this mapping process, a specific regularization term is used to satisfy a crucial property: each filter of the concept encoder should learn the visual feature corresponding to only one concept $c^k$. Let the output of the concept encoder $\psi$ be the set of features $\textbf{v} = \{v_1^i,...,v_k^i\}$, which result from a GAP operation over each feature map $\mathcal{A}_k^{i}$. To relate each visual feature $v_k^i$ to the presence or absence of a particular concept, the vector $\textbf{z} = \{z_1^i, ..., z_k^i\}$ was defined to represents the presence ($z_k^i = 1$) or absence ($z_k^i = 0$) of a concept $c^k$ in the image $x^{i}$. To encourage each filter of the concept encoder $\psi$ to be activated only when the concept $c^k$ is present in the image, we compare the ground-truth concept $z_k^i$ with the estimated concept $\sigma(v_k^i)$, where $\sigma$ is the sigmoid activation function, by minimizing the binary cross-entropy loss:

\begin{equation}
\label{eq:uniqueness}
    \mathcal{L}_u = \sum_k -(z_k^i \log(\sigma(v_k^i)) + (1-z_k^i) \log(1-\sigma(v_k^i)).
\end{equation}


\paragraph{Mapping Consistency Loss}


Assuming that an image can have more than one associated concept, it is crucial to isolate how a filter is activated when a concept is present. As such, the visual features of the feature map $\mathcal{A}_k^{i}$ and the respective word phrase $s_k$ are mapped to a joint embedding space using two embedding matrices: i the visual feature embedding matrix $\mathcal{E}_v \in \mathbb{R}^{w \times h \times d_e}$ and (ii) the word phrase embedding matrix $\mathcal{E}_s \in \mathbb{R}^{d \times d_e}$, where $d_e$ is the dimension of the embedding layer. Then, an embedded visual feature is expressed as $\tilde{v}_k^{i} = \mathcal{E}_v . \mathcal{A}_k^{i}$ and an embedded word phrase is defined as $\tilde{s}_k = \mathcal{E}_s . s_k$. Therefore, the similarity between the visual feature $\tilde{v}_k^{i}$ and its corresponding word phrase $\tilde{s}_k$ should be higher than the visual feature $\tilde{v}_k^{i}$ and any other word phrase $\tilde{s}_{k' \neq k}$. This is achieved by the triplet loss, defined as follows:
\begin{equation}
\label{eq:semantic_loss}
    \mathcal{L}_s = \sum_k z_k^i \sum_{k' \neq k} \max (0, \frac{\tilde{v}_k^i . \tilde{s}_{k'}}{\left\|\tilde{v}_k^i\right\|\left\|\tilde{s}_{k'}\right\|} - \frac{\tilde{v}_k^i . \tilde{s}_k}{\left\|\tilde{v}_k^i\right\|\left\|\tilde{s}_k\right\|} + \alpha),
\end{equation}
where $\tilde{v}_k^i$ is the embedded visual feature, $\tilde{s}_k$ is the embedded word phrase, and $\alpha$ is a hyperparameter denoting a margin between positive and negative pairs. In order to further improve the mapping consistency and to ensure that each filter isolates some concept, for each concept $k$ that is in image $x^{i}$ but not in image $x^{i'}$ its word phrase $\tilde{s}_k$ should be more similar to the visual feature $\tilde{v}_k^i$ than to the visual feature of the counter image $x^{i'}$. This rationale penalizes the cases where the word phrase $\tilde{s}_k$ is closer to the visual features of an image $x^{i'}$ that does not have the concept $c^k$ than that of an image $x^{i}$ that has the concept $c^k$. This constraint is achieved by minimizing the following loss function:
\begin{equation}
\label{eq:counter_loss}
  \mathcal{L}_d = \sum_{z_k^{i'} - z_k^{i} > 0}  \max (0, \frac{\tilde{v}_k^{i'} . \tilde{s}_k}{\left\|\tilde{v}_k^{i'}\right\|\left\|\tilde{s}_k\right\|} - \frac{\tilde{v}_k^i . \tilde{s}_k}{\left\|\tilde{v}_k^i\right\|\left\|\tilde{s}_k\right\|} + \beta),
\end{equation}
where $\beta$ is a hyperparameter.

Then, the mapping consistency loss is defined as the sum of the equation (\ref{eq:semantic_loss}) and equation (\ref{eq:counter_loss}): $\mathcal{L}_m = \mathcal{L}_s +\mathcal{L}_d$.



\paragraph{Coherence Loss}

To encourage the activations of each filter of the concept encoder to concentrate on the lesion region when the concept is present in the image and to penalize the activations when the concept is absent in the image, we include an additional regularization term, denoted as Coherence Loss $\mathcal{L}_c$. This regularization is achieved by modifying the Dice-Sørensen coefficient~\cite{dice1945measures} to measure the divergence between the feature map $\mathcal{A}_k^{i}$ of the concept encoder $\psi$ and the object segmentation mask $\mathcal{M}_k^{i}$, expressed as follows:
\begin{equation}
    \mathcal{L}_c = \sum_k 1 - \frac{2 \sum \mathcal{A}_k^{i} \mathcal{M}_k^{i}}{\sum {\mathcal{A}_k^{i}}^2 \sum {\mathcal{M}_k^{i}}^2 + \epsilon},
\end{equation}
where $\epsilon$ is a parameter to avoid division by $0$.

Taking into consideration all the above-described combined losses, and adding the cross-entropy loss as the classification loss $\mathcal{L}_A(\hat{y}, y)$, the overall objective function can be written as follows:
\begin{equation}
    \mathcal{L} = \mathcal{L}_A + \lambda(\mathcal{L}_u + \mathcal{L}_m) + \gamma \mathcal{L}_c,
\end{equation}
where $\lambda \in [0,1]$ is the hyperparameter weighting the influence of the concept uniqueness loss $\mathcal{L}_u$ and the mapping consistency loss $\mathcal{L}_m$ in the total loss $\mathcal{L}$, and $\gamma \in [0,1]$ is the hyperparameter that controls the effect of the coherence loss $\mathcal{L}_c$. 

\subsection{Contribution to Classification Decision}

As previously mentioned in the introductory part of section \ref{sec:method}, it is desirable to decompose the final classification decision in terms of human interpretable concepts along with their respective contribution to the model decision. Since the weight matrix ${\mathbf{W}} \in \mathbb{R}^{\left| C \right| \times n}$ of the FC layer of the classifier denotes the importance of each concept to the final class prediction, we can leverage this information to calculate the contribution of each concept $c^k$ to the final decision. This is achieved by multiplying the visual feature $v_k^i$ with the corresponding weight $w_{k,c}$ of the FC layer. The result of this operation is passed through a softmax layer to obtain a probability estimation. With these concepts contributions, we can construct more informative and coherent textual explanations (Figure \ref{fig:method}) than relying on other text-based approaches that have some limitations in generating coherent and structured text reports. Furthermore, the final prediction is directly extrapolated from the weighted sum of the concepts contributions, expressed as $\hat{y} = {\arg\max}(\textbf{W}^T . \textbf{v})$.



\subsection{Segmentation Module}
\label{sec:segmentation}

We propose using segmentation masks as a hard attention mechanism in our method. Its primary purpose is to improve the model intepretability while expecting that the model ignores artifacts that could be present in the image. The segmentation masks are also used to guide the convolutional filters of the concept encoder $\psi$ to be activated only in relevant regions (coherence loss), i.e., the lesion region, resulting in more coherent and refined learned concepts. To this end, we used the ground-truth skin segmentation masks provided in the HAM10000~\cite{HAM10000} dataset for the segmentation task, and we adopted the DeepLabV3~\cite{deeplabv3} model, a state-of-the-art semantic segmentation model that achieves good results in a panoply of datasets. After training, the skin image is fed to the model to obtain the segmentation masks. Then, with the masks and the skin images, we use a \textit{bitwise-and} to remove all non-relevant pixels, and the produced result is fed to the next stage of the model, as depicted in Figure \ref{fig:method}. This way, we are not dependent on additional annotations of ground-truth segmentation masks, as our trained segmentation model can generate accurate segmentation masks for the new data.

\section{Experiments}
\label{sec:experiments}


\subsection{Experimental Setup}

\paragraph{Datasets.} 
The supervised constraint of our method in concepts annotations limited our selection of datasets to PH$^2$~\cite{PH2} and Derm7pt~\cite{DERM7PT}, as both datasets comprise ground-truth annotations regarding the presence or absence of several dermoscopic criteria. PH$^2$ dataset contains $200$ dermoscopic images of melanocytic lesions, including ``common nevi'', ``atypical nevi'', and ``melanoma''. Additionally, PH$^2$ includes ground-truth lesion segmentation masks. Derm7pt comprises over $2,000$ clinical and dermoscopic  images, which we filtered to obtain $827$ images of ``nevus'' and ``melanoma'' classes. Although Derm7pt does not provide ground-truth segmentation masks, we performed its manual annotation using the Computer Vision Annotation Tool (CVAT). To take advantage of both datasets jointly, we combined the ``common nevi'' and ``atypical nevi'' classes of the PH$^2$ into one global class label denominated ``Nevus'' to construct a largest dataset, named PH2D7, comprising $1,027$ clinical and dermoscopy images, converting our classification task into a binary classification problem (``Nevus'' vs ``Melanoma''). 

\paragraph{Dermoscopic Concepts.} Both PH$^2$~\cite{PH2} and Derm7pt~\cite{DERM7PT} datasets comprise annotations of several dermoscopic criteria, which we refer to ``dermoscopic concepts''. Specifically, we considered $8$ dermoscopic concepts $\textbf{c}$ according to the annotated dermoscopic criteria: ``Atypical Pigment Network'' (APN), ``Typical Pigment Network'' (TPN), ``Blue Whitish-Veil'' (BWV), ``Irregular Streaks'' (ISTR), ``Regular Streaks'' (RSTR), ``Regular Dots and Globules'' (RDG), ``Irregular Dots and Globules'' (IDG) and ``Regression Structures'' (RS).


\paragraph{Implementation Details.} The training of the proposed framework proceeds in three stages: i training only the concept encoder $\psi$ and the classifier; (ii) training the whole network, and (iii) training only the classifier for a few more epochs. We use ResNet-101~\cite{he2016deep}, DenseNet-201~\cite{huang2017densely} and SEResNeXt~\cite{hu2018squeeze} architectures pretrained on
ImageNet~\cite{deng2009imagenet} as our feature extractor $\phi$. The output of the feature extractor $\phi$ is passed to the concept encoder $\psi$ to output the visual features $\textbf{v} = \{v_1^i, ..., v_k^i\}$. The classifier is a single fully connected layer that receives the visual features $\textbf{v}$ of the concept encoder $\psi$ and predicts the class label $\hat{y}$. We adopt the Adam~\cite{kingma2014adam} optimizer during training and set $\lambda = 0.4$, $\alpha = 1.0$, and $\beta = 0.5$ in our experiments. Regarding the $\gamma$ hyperparameter, we fine-tuned its value according to the model and dataset used. All details and PyTorch code are available at \url{https://github.com/CristianoPatricio/coherent-cbe-skin}.

\subsection{Quantitative Evaluation}
\label{subsec:qualitative_evaluation}

\begin{table*}
  \centering
  \setlength{\tabcolsep}{7pt}
  \resizebox{\textwidth}{!}{%
  \begin{tabular}{clccccccccc}
    \toprule
    \multirow{3}{*}{\textbf{Preprocessing}} & \multirow{3}{*}{\textbf{Method}} & \multicolumn{3}{c}{\textbf{SEResNeXt}} & \multicolumn{3}{c}{\textbf{DenseNet-201}} & \multicolumn{3}{c}{\textbf{ResNet-101}} \\
    \cmidrule{3-5} \cmidrule{6-8} \cmidrule{9-11}
    & & PH$^2$ & D7 & PH2D7 & PH$^2$ & D7 & PH2D7 & PH$^2$ & D7 & PH2D7\\
    & & \#25 / \#5 & \#219 / \#101 & \#239 / \#106 & \#25 / \#5 & \#219 / \#101 & \#239 / \#106 & \#25 / \#5 & \#219 / \#101 & \#239 / \#106 \\
    \midrule
    \midrule
    \multirow{4}{*}{Raw} 
    & Black-box~\cite{lopez2017skin} & 84.00 $\pm$ 3.27   & 73.13 $\pm$ 1.42          & 74.88 $\pm$ 2.72          & 85.33 $\pm$ 1.89  & 79.90 $\pm$ 0.74          & 80.48 $\pm$ 0.60          & 80.00 $\pm$ 3.27 & 76.15 $\pm$ 0.39          & 77.58 $\pm$ 1.09 \\
    & Baseline~\cite{wickramanayake2021comprehensible}  & 92.00 $\pm$ 3.27   & 80.00 $\pm$ 0.92         & 80.10 $\pm$ 0.36          & {\textbf{94.67 $\pm$ 1.89}}  & 81.72 $\pm$ 0.47          & {\textbf{83.04 $\pm$ 0.72}}        & 96.00 $\pm$ 0.00 & 81.77 $\pm$ 1.03          & {\textbf{83.38 $\pm$ 0.60 }}\\
    & ExAID~\cite{lucieri2022exaid}     & -       & {\textbf{83.60}}          & -              & -      & -              & -              & -     & -              & -     \\
    & \textbf{Ours}      & \textbf{94.67 $\pm$ 1.89}   & 80.42 $\pm$ 0.53          & \textbf{80.77 $\pm$ 0.49 }         & 94.00 $\pm$ 2.00  & \textbf{82.03 $\pm$ 0.47 }         & 82.90 $\pm$ 1.03          & 96.00 $\pm$ 0.00 & \textbf{82.60 $\pm$ 0.39 }         & 82.51 $\pm$ 0.90 \\
    \midrule
    \multirow{3}{*}{DLV3} 
    & Black-box~\cite{lopez2017skin} & 90.67 $\pm$ 1.89  & 76.15 $\pm$ 1.41          & 79.23 $\pm$ 1.81         & 94.67 $\pm$ 1.89  & 78.96 $\pm$ 0.53         & 78.65 $\pm$ 1.21         & 86.67 $\pm$ 1.89 & 79.58 $\pm$ 0.53         & 80.77 $\pm$ 1.54 \\
    & Baseline~\cite{wickramanayake2021comprehensible}  & 93.33 $\pm$ 1.89  & {\textbf{84.27 $\pm$ 0.78}}          & 83.96 $\pm$ 0.14          & 96.00 $\pm$ 0.00  & {\textbf{85.94 $\pm$ 0.44}}       & 85.51 $\pm$ 0.00        & 96.00 $\pm$ 0.00 & 81.67 $\pm$ 1.03          & {\textbf{84.44 $\pm$ 0.36}} \\
    & \textbf{Ours}      & \textbf{96.00 $\pm$ 0.00}  & 84.06 $\pm$ 0.44 & \textbf{84.44 $\pm$ 0.55} & 96.00 $\pm$ 0.00  & {85.73 $\pm$ 0.29} & \textbf{85.89 $\pm$ 0.27} & 96.00 $\pm$ 0.00 & \textbf{83.75 $\pm$ 0.26} & 84.15 $\pm$ 0.14\\
    \midrule
    \multirow{3}{*}{Manually} 
    & Black-box~\cite{lopez2017skin} & 90.67 $\pm$ 4.99   & 75.93 $\pm$ 0.88          & 77.97 $\pm$ 0.63          & 90.67 $\pm$ 1.89 & 77.71 $\pm$ 0.53          & 79.32 $\pm$ 2.12          & 93.33 $\pm$ 1.89 & 75.00 $\pm$ 0.68          & 75.94 $\pm$  0.85\\
    & Baseline~\cite{wickramanayake2021comprehensible}  & 96.00 $\pm$ 0.00   & {\textbf{84.17 $\pm$ 0.39   }}       & 83.48 $\pm$ 0.95          & 85.33 $\pm$ 7.54  & 82.92 $\pm$ 0.53  & 82.75 $\pm$ 0.72         & 96.00 $\pm$ 0.00 & {\textbf{82.66 $\pm$ 0.47}}           & 74.30 $\pm$ 3.62 \\
    & \textbf{Ours}      & 96.00 $\pm$ 0.00   & 83.02 $\pm$ 0.39          & \textbf{84.54 $\pm$ 0.49  }        & \textbf{93.33 $\pm$ 1.89}  & \textbf{83.58 $\pm$ 0.24}          & \textbf{84.35 $\pm$ 0.85}         & 96.00 $\pm$ 0.00 & 82.29 $\pm$ 1.85          & \textbf{83.19 $\pm$ 0.29} \\
    \bottomrule
  \end{tabular}%
  }
  \caption{\textbf{Quantitivate results}. Quantitative comparison results (accuracy in \%) of different methods under three preprocessing strategies on PH$^2$, Derm7pt, and PH2D7 datasets. ``\textit{Raw}'' indicates that no segmentation was applied in the input images; ``\textit{DLV3}'' denotes that images were segmented using the segmentation masks provided by the trained DeepLabV3 on HAM10000 dataset, and ``\textit{Manually}'' denotes the images segmented with the ground-truth segmentation masks provided by the datasets. The number of images (\#Nevus / \#Melanoma) considered at evaluation are given below the dataset. The results were obtained in three runs under different seeds. \textbf{Bold} highlights the best results.}
  \label{tab:global_results}
\end{table*}

In Table \ref{tab:global_results}, we report the classification accuracy (in \%) of our method (\textbf{Ours}) and compare it with other methods under three preprocessing strategies on the considered datasets (PH$^2$, Derm7pt and PH2D7) for each adopted feature extractor architecture (ResNet-101, DenseNet-201 and SEResNeXt). Specifically, we compare the performance of our method (\textbf{Ours}) with the baseline model ~\cite{wickramanayake2021comprehensible} and a standard deep learning-based approach~\cite{lopez2017skin} which we refer to as ``Black-box''. From the reported results in Table \ref{tab:global_results}, we can observe that our approach (\textbf{Ours}) outperforms the black-box models by a large margin and performs comparably with the baseline models, even surpassing its performance in some settings and datasets, confirming that the $\mathcal{L}_c$ helps improve the performance. Another important observation is that we achieved the best classification accuracy when adopting the segmentation strategy, confirming the utility of the segmentation masks for improving the final classification performance. Furthermore, using the segmentation masks provided by DeepLabV3, we obtain the best results compared to the values when using
the ground-truth (Manually) segmentation masks. Note that our method also achieves superior performance, considering the best setting, compared to the recently proposed ExAID~\cite{lucieri2022exaid} framework ($84.06\%$ vs $83.60 \%$), that despite being a concept-based approach, does not follow the philosophy of the inherently interpretable models. In addition, the discrepancy between the results obtained by the PH $^2$ dataset and the Derm7pt dataset could be related to the type of images each dataset contains.  Although the PH$^2$ is a small dataset, it contains only dermoscopic images, whereas Derm7pt contains both dermoscopic and clinical images, which ultimately impacts the performance of our model when evaluated on the Derm7pt dataset. We also report the sensitivity, specificity, and the area under the receiver operating characteristic curve (AUC) in Table \ref{tab:se_sp_auc}. The obtained results allow us to conclude that the detection of ``Melanoma'' is more challenging in the case of the Derm7pt and the PH2D7 datasets (a lower value of sensitivity). This fact can be explained due to the less representativity of class ``Melanoma'' in these datasets. On the other hand, we observe a superior performance when DenseNet-201 is adopted as feature extractor in our proposed method, which was also confirmed by the results in Table \ref{tab:global_results}.

\begin{table}[!htbp]
  \centering
  \setlength{\tabcolsep}{6pt}
  \resizebox{.35\textwidth}{!}{%
  \begin{tabular}{lrccc}
    \toprule
    \textbf{Model} & \textbf{Dataset} & \textbf{Sens.} & \textbf{Spec.} & \textbf{AUC} \\
    \midrule
    \midrule
    \multirow{3}{*}{SEResNext} 
    & PH$^2$ & 1.000 & 0.950 & 0.975 \\
    & D7 & 0.535 & 0.977 & 0.756 \\
    & PH2D7 & 0.594 & 0.962 & 0.778 \\
    \midrule
    \multirow{3}{*}{DenseNet-201} 
    & PH$^2$ & 1.000 & 0.950 & {0.975}\\
    & D7 & 0.574 & 0.973 & \textbf{0.773} \\
    & PH2D7 & 0.623 & 0.967 & \textbf{0.795} \\
    \midrule
    \multirow{3}{*}{ResNet-101} 
    & PH$^2$ & 1.000 & 0.950 & 0.975 \\
    & D7 & 0.564 & 0.968 & 0.766 \\
    & PH2D7 & 0.594 & 0.950 & 0.772 \\
    \bottomrule
  \end{tabular}%
  }
  \caption{\textbf{Comparison of different feature extractors.} The classification score of the proposed method was evaluated with different architectures and using the DLV3 images of PH$^2$, Derm7pt, and PH2Derm7pt. ``Sens.'' denotes Sensitivity, ``Spec.'' denotes Specificity.}
  \label{tab:se_sp_auc}
\end{table}

\paragraph{Concepts Prediction}

We evaluate the predictive performance of our concept encoder in estimating the presence or absence of dermoscopic concepts, considering the best preprocessing strategy, i.e. DLV3. As denoted in equation (\ref{eq:uniqueness}), the result of a GAP operation over the feature maps $\mathcal{A}^i$ allows mapping the spatial information to a scalar value denoting the presence or absence of a concept. Based on this, and after passing the results through a hyperbolic tangent (tanh) layer, we can infer the presence of the concept. Table \ref{tab:f1_concepts} reports the predictive performance of our concept encoder in terms of the F1-score. We compare the performance of our proposed method (\textbf{Ours}) with the baseline~\cite{wickramanayake2021comprehensible}. The results indicate that the predictive performance consistently improves compared to the results of the baseline method. We also report in Table \ref{tab:f1_concepts} the concept explanation error ($L_2$) to measure how close are the predicted concepts to the ground-truth concepts. Specifically, we calculate the $L_2$ distance between the predicted and ground-truth concepts. From the results of Table \ref{tab:f1_concepts}, we can observe that the minimum value is achieved when predicting concepts on the PH$^2$ dataset ($< 8.00$). Also, the performance also (lower the better) when using our method (\textbf{Ours}).

\begin{table}[!htbp]
  \centering
  \setlength{\tabcolsep}{5pt}
  \resizebox{.35\textwidth}{!}{%
  \begin{tabular}{lrcccc}
    \toprule
    \multirow{2}{*}{\textbf{Model}} & \multirow{2}{*}{\textbf{Dataset}} & \multicolumn{2}{c}{\textbf{Baseline}} & \multicolumn{2}{c}{\textbf{Ours}} \\
    \cmidrule{3-4} \cmidrule{5-6}
    & & F1 $\uparrow$ & $L_2$ $\downarrow$ & F1 $\uparrow$ & $L_2$ $\downarrow$\\
    \midrule
    \midrule
    \multirow{3}{*}{SEResNext} 
    & PH$^2$ & 0.67 & 7.87 & \textbf{0.69} & \textbf{7.55} \\
    & D7 & 0.70 & 27.57 & \textbf{0.71} & \textbf{25.24} \\
    & PH2D7 & 0.68 & 27.82 & \textbf{0.69} & \textbf{27.17} \\
    \midrule
    \multirow{3}{*}{DenseNet-201} 
    & PH$^2$ & 0.57 & 7.68 &  \textbf{0.68} & 7.68 \\
    & D7 & 0.00 & 33.36 & \textbf{0.71} & \textbf{25.67} \\
    & PH2D7 & 0.00 & 34.63 & \textbf{0.70} & \textbf{27.53}\\
    \midrule
    \multirow{3}{*}{ResNet-101} 
    & PH$^2$ & \textbf{0.72} & 7.48  & 0.70 & \textbf{7.35} \\
    & D7 & 0.66 & 26.12 & \textbf{0.71} & \textbf{25.85} \\
    & PH2D7 & 0.68 & 27.64 & \textbf{0.69} & \textbf{27.68} \\
    \bottomrule
  \end{tabular}%
  }
  \caption{\textbf{Comparison of concepts predictions}. Performance comparison of concept prediction (F1-score) and concept explanation error ($L_2$ distance) for different feature extractors on PH$^2$, Derm7pt and PH2D7 datasets.}
  \label{tab:f1_concepts}
\end{table}

\subsection{Qualitative Evaluation}

\begin{figure*}[!htbp]
    \centering
    \includegraphics[width=0.95\textwidth]{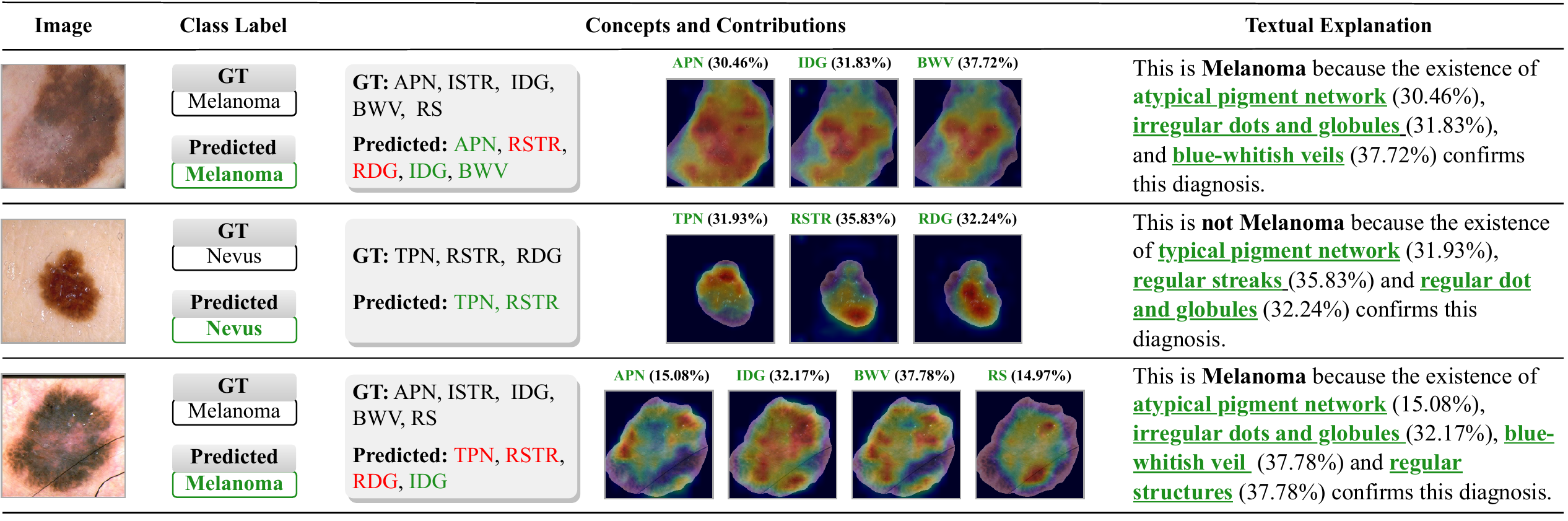}
    \caption{\textbf{Qualitative results}. Examples of provided explanations by our method given skin images of different class labels. Besides the predicted class label, our model can provide individual concept interpretation and a textual explanation that reflects the estimated concepts and their contribution (in \%) to the final class decision.}
    \label{fig:qualitative_eval}
\end{figure*}

Figure \ref{fig:qualitative_eval} shows examples of ``Nevus'' and ``Melanoma'' cases predicted by our model. As observed in the first example (first line of Figure \ref{fig:qualitative_eval}), the image was correctly classified as ``Melanoma''. However, two predicted concepts were incorrectly estimated (RSTR and RDG). After calculating the contribution (in \%) of each concept to the decision, only positive contributions were given to the correctly predicted concepts, i.e. APN, IDG and BWV. The contribution of the concepts RSTR and RDG was null, since the weight of the FC layer associated with these concepts reduced their importance. As such, we can view this concept contribution estimation as a second opinion to the predicted concepts by the concept encoder. Similar conclusions are applicable to the third example of Figure \ref{fig:qualitative_eval}. This way, since the textual explanation relies only on concepts with a positive contribution, the final decision is sustained by valid arguments and correctly identified concepts. Regarding the visualization of the learned concepts by the concept encoder, we can observe that the highlighted regions coincide with the lesion region, evidencing critical areas for the detection of melanoma, and which are clinically coherent with the common signs of melanoma according to the ABCD rule~\cite{Nachbar1994JAAD}: asymmetry, border irregularity, multi-color and dermoscopic structure. In the second example (second line of Figure \ref{fig:qualitative_eval}), the image was correctly classified as ``Nevus'' and the estimated concepts by the concept encoder were also predicted correctly, except the RDG concept. Nevertheless, this concept is included in the textual justification for having a positive contribution to the final decision.



\paragraph{Concepts Localization Coherence}

To assess whether the proposed strategy provides consistent concept locations along different CNN architectures, we measured the correlation between the activations of different architectures. For this, we determined the highly activated regions of each concept by thresholding  its activation map (based on the activations distribution of training data). The binary maps obtained from each model were compared using the Dice coefficient in a pairwise manner. Table \ref{tab:cos_concepts} reports the average dice similarity coefficient (DSC) between the binary concept maps predicted by the proposed model using three feature extractors (ResNet-101, DenseNet-201 and SEResNext). The concepts are consistent among the different model architectures since the Dice coefficient is relatively close to the perfect value, i.e. $1.0$. The lack of results for the remaining concepts is due to the disagreement between the different models in correctly predicting those concepts. Additionally, we also measured the correlations between the feature representations of each highly activated region along different architectures. Specifically, we calculated the average cosine similarity between the features representation of each set of concept patches of the same concept, using the feature extractors mentioned above. The results on test set are reported in Table \ref{tab:cos_concepts}, and indicate a high similarity between the concepts, confirming the consistency of the concept encoder in activating meaningful concept locations. The omission of results for some concepts, namely ISTR and RS, is due to the less representativity of these concepts in the data, causing incorrect predictions, i.e., a null contribution of the concept to the prediction.

\begin{table}[!htbp]
  \centering
  \setlength{\tabcolsep}{6pt}
  \resizebox{.35\textwidth}{!}{%
  \begin{tabular}{lcccc}
    \toprule
    \multirow{2}{*}{\textbf{Concept}} & \multicolumn{3}{c}{\textbf{Cosine Similarity}}  & \multirow{2}{*}{\textbf{DSC}} \\
    \cmidrule{2-4}
    & ResNet & DenseNet & SEResNeXt & \\
    \midrule
    \midrule
    TPN & 0.92 & 0.86 & 0.89 & \cellcolor{lavender} 0.78\\
    BWV & 0.89 & - & - & \cellcolor{lavender} -\\
    APN & 0.85 & 0.84 & 0.83 & \cellcolor{lavender} 0.80\\
    RSTR & 0.86 & 0.83 & 0.84 & \cellcolor{lavender} 0.91\\
    RDG & 0.92 & 0.83 & 0.85 & \cellcolor{lavender} 0.81\\
    IDG & 0.95 & 1.00 & 0.89 & \cellcolor{lavender} -\\
    \bottomrule
  \end{tabular}%
  }
  \caption{\textbf{Concepts location consistency}. Cosine similarity between the feature representations of the concept patches for each dermoscopic concept. The values colored with a gray background denotes the average dice similarity coefficient (DSC) between the concept patches generated by different model variants. A value of 1.0 represents a perfect alignment.}
  \label{tab:cos_concepts}
\end{table}

\subsection{Ablation Studies}
\label{subsec:ablation}



\paragraph{Importance of Segmentation Masks}

\begin{figure}
    \centering
    \includegraphics[width=0.35\textwidth]{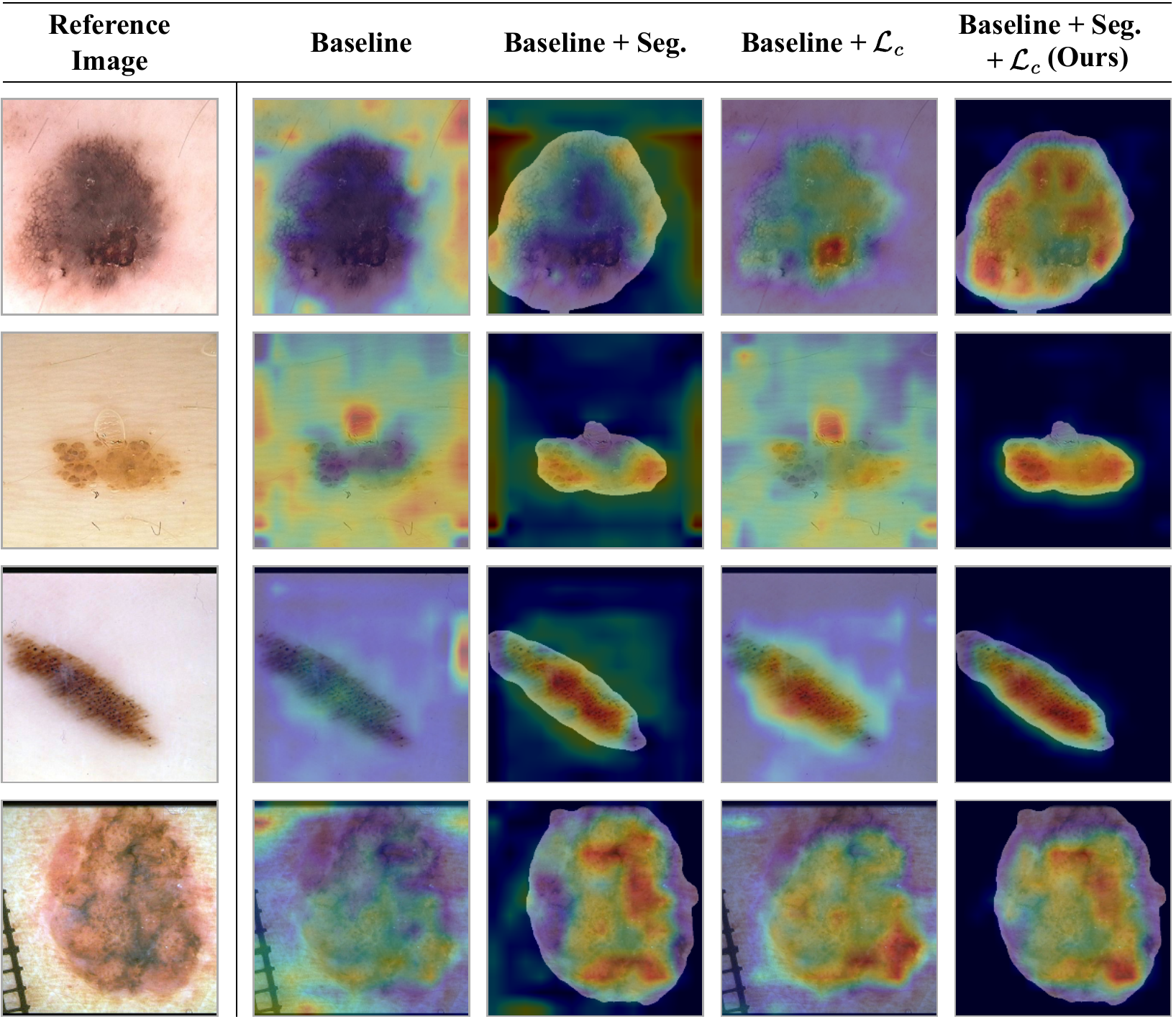}
    \caption{\textbf{Effect of segmentation}. Examples of generated feature maps by the concept encoder showing the improvement in isolating a concept inside the lesion region as the segmentation strategy and coherence loss are added to the baseline model.}
    \label{fig:ablation_segmented}
\end{figure}

For studying the impact of the segmentation module, we ablated this preprocessing step from our approach by relying on the raw images during training. The results are provided in Table \ref{tab:global_results} and show a clear improvement (Raw vs DLV3) when using this preprocessing strategy, confirming our insight that the segmentation module is indeed a hard attention mechanism to prevent the model from giving attention to some artifacts that could impact the model performance. Additionally, we compared the visual coherence of the feature maps produced by the concept encoder (Figure \ref{fig:ablation_segmented}), and it can be observed that the use of the segmentation module decreases the attention given to the background, but does not completely focus on the region of the lesion. However, its combination with the coherence loss the attention concentrates solely on the lesion region.


\paragraph{Effect of Regularization Components}

To examine the effectiveness of each regularization term in the global objective function, we have incrementally added to the classification loss $\mathcal{L}_A$ each of the loss components to perceive its effect on the global classification performance. From the reported results in Table \ref{tab:ablation_losses}, we conclude that by adding each of the regularization terms described in section \ref{sec:loss}, the global classification performance is not significantly affected by the imposed constraints. Note that the best performance configuration is achieved when all loss components are jointly considered, accounting for accuracy and interpretability simultaneously (see Figure \ref{fig:ablation_segmented}). Compared to the simplest configuration, which only uses the $\mathcal{L}_A$, our considered objective loss (last configuration of Table \ref{tab:ablation_losses}) does not significantly impact the overall performance.

\begin{table}[!h]
  \centering
  \setlength{\tabcolsep}{7pt}
  \resizebox{.4\textwidth}{!}{%
  \begin{tabular}{lccc}
    \toprule
    \textbf{Method} & PH$^2$ & D7 & PH2D7\\
    \midrule
    \midrule
    $\mathcal{L}_A$ & 94.67 $\pm$ 1.89  & 85.63 $\pm$ 0.26 & 85.99 $\pm$ 0.76 \\
    $\mathcal{L}_A + \mathcal{L}_u$ & 96.00 $\pm$ 0.00 & 86.25 $\pm$ 0.51 & 86.38 $\pm$ 0.24 \\
    $\mathcal{L}_A + \mathcal{L}_m$ & 94.67 $\pm$ 1.89 & 85.00 $\pm$ 0.68 & 86.38 $\pm$ 0.41 \\
    $\mathcal{L}_A + \mathcal{L}_c$ & 96.00 $\pm$ 0.00 & 85.42 $\pm$ 0.29 & 85.51 $\pm$ 0.41 \\
    $\mathcal{L}_A  + \mathcal{L}_m   + \mathcal{L}_u$ & 94.67 $\pm$ 1.89  & {86.15 $\pm$ 0.15}       & 85.41 $\pm$ 0.14 \\
    $\mathcal{L}_A  + \mathcal{L}_u   + \mathcal{L}_c$ & 96.00 $\pm$ 0.00  & 85.52 $\pm$ 0.39  & 86.09 $\pm$ 0.24 \\
    $\mathcal{L}_A + \mathcal{L}_m   + \mathcal{L}_u + \mathcal{L}_c$ & 96.00 $\pm$ 0.00 & 85.73 $\pm$ 0.29 & {85.89 $\pm$ 0.27} \\
    \bottomrule
  \end{tabular}%
  }
  \caption{\textbf{Effect of coherence loss}. Comparison of classification accuracy given different variants of the global objective function using the DenseNet-101 as feature extractor. The results were obtained in three runs under different seeds.}
  \label{tab:ablation_losses}
\end{table}

\section{Conclusions and Future Work}
\label{sec:conclusions}
This paper introduces an inherently interpretable concept-based model for skin lesion diagnosis. Incorporating the segmentation strategy as a hard attention mechanism and the coherence loss term into a general concept-based approach assures the visual coherence of concept activations by the concept encoder whilst enhancing individual concept interpretation and improving the overall classification accuracy of skin lesion diagnosis. Because of the constraints imposed by the interpretable internal architecture, our experiments demonstrated that our model performed comparably and in many cases was even superior to the black-box and baseline competitors. However, our approach is dependent on concepts annotations. In future work, we intend to adopt self-supervised techniques to mitigate this dependency. Nevertheless, we hope our method will be helpful for further investigation in inherently interpretable models and to encourage the trustworthiness and acceptance of automated diagnosis systems in daily clinical routines. 

\section*{Acknowledgements}
This work was funded by the Portuguese Foundation for Science and Technology (FCT) under the PhD grant ``2022.11566.BD'', and supported by NOVA LINCS (UIDB/04516/2020) with the financial support of FCT.IP.

{\small
\bibliographystyle{ieee_fullname}
\bibliography{paper}

\begin{thebibliography}{10}\itemsep=-1pt

\bibitem{adebayo2018sanity}
Julius Adebayo, Justin Gilmer, Michael Muelly, Ian Goodfellow, Moritz Hardt,
  and Been Kim.
\newblock {Sanity Checks for Saliency Maps}.
\newblock In {\em Proceedings of International Conference on Neural Information
  Processing Systems (NeurIPS)}, 2018.

\bibitem{argenziano1998epiluminescence}
Giuseppe Argenziano, Gabriella Fabbrocini, Paolo Carli, Vincenzo De~Giorgi,
  Elena Sammarco, and Mario Delfino.
\newblock {Epiluminescence Microscopy for the Diagnosis of Doubtful Melanocytic
  Skin Lesions: Comparison of the ABCD Rule of Dermatoscopy and a new 7-point
  Checklist based on Pattern Analysis}.
\newblock {\em Archives of Dermatology}, 134(12):1563--1570, 1998.

\bibitem{barata2021explainable}
Catarina Barata, M.~Emre Celebi, and Jorge~S. Marques.
\newblock {Explainable Skin Lesion Diagnosis Using Taxonomies}.
\newblock {\em Pattern Recognition}, 110:107413, 2021.

\bibitem{deeplabv3}
Liang-Chieh Chen, George Papandreou, Florian Schroff, and Hartwig Adam.
\newblock {Rethinking Atrous Convolution for Semantic Image Segmentation}.
\newblock {\em arXiv preprint arXiv:1706.05587}, 2017.

\bibitem{chen_2020_NMI}
Zhi Chen, Yijie Bei, and Cynthia Rudin.
\newblock {Concept Whitening for Interpretable Image Recognition}.
\newblock {\em Nature Machine Intelligence}, 2(12):772--782, 2020.

\bibitem{deng2009imagenet}
Jia Deng, Wei Dong, Richard Socher, Li-Jia Li, Kai Li, and Li Fei-Fei.
\newblock {ImageNet: A Large-Scale Hierarchical Image Database}.
\newblock In {\em Proceedings of the IEEE Conference on Computer Vision and
  Pattern Recognition (CVPR)}, pages 248--255, 2009.

\bibitem{dice1945measures}
Lee~R Dice.
\newblock {Measures of the Amount of Ecologic Association Between Species}.
\newblock {\em Ecology}, 26(3):297--302, 1945.

\bibitem{esteva2017dermatologist}
Andre Esteva, Brett Kuprel, Roberto~A Novoa, Justin Ko, Susan~M Swetter,
  Helen~M Blau, and Sebastian Thrun.
\newblock {Dermatologist-level classification of skin cancer with deep neural
  networks}.
\newblock {\em Nature}, 542(7639):115--118, 2017.

\bibitem{Fang_MM2020}
Zhengqing Fang, Kun Kuang, Yuxiao Lin, Fei Wu, and Yu-Feng Yao.
\newblock {Concept-based Explanation for Fine-grained Images and Its
  Application in Infectious Keratitis Classification}.
\newblock In {\em Proceedings of the ACM International Conference on
  Multimedia}, pages 700--708, 2020.

\bibitem{Graziani_2018}
Mara Graziani, Vincent Andrearczyk, and Henning M{\"u}ller.
\newblock Regression concept vectors for bidirectional explanations in
  histopathology.
\newblock In {\em Understanding and Interpreting Machine Learning in Medical
  Image Computing Applications}, pages 124--132. 2018.

\bibitem{he2016deep}
Kaiming He, Xiangyu Zhang, Shaoqing Ren, and Jian Sun.
\newblock {Deep Residual Learning for Image Recognition}.
\newblock In {\em Proceedings of the IEEE Conference on Computer Vision and
  Pattern Recognition (CVPR)}, pages 770--778, 2016.

\bibitem{hu2018squeeze}
Jie Hu, Li Shen, and Gang Sun.
\newblock {Squeeze-and-Excitation Networks}.
\newblock In {\em Proceedings of the IEEE Conference on Computer Vision and
  Pattern Recognition (CVPR)}, pages 7132--7141, 2018.

\bibitem{huang2017densely}
Gao Huang, Zhuang Liu, Laurens Van Der~Maaten, and Kilian~Q Weinberger.
\newblock {Densely Connected Convolutional Networks}.
\newblock In {\em Proceedings of the IEEE Conference on Computer Vision and
  Pattern Recognition (CVPR)}, pages 4700--4708, 2017.

\bibitem{DERM7PT}
Jeremy Kawahara, Sara Daneshvar, Giuseppe Argenziano, and Ghassan Hamarneh.
\newblock {Seven-Point Checklist and Skin Lesion Classification Using Multitask
  Multimodal Neural Nets}.
\newblock {\em IEEE Journal of Biomedical and Health Informatics},
  23(2):538--546, 2019.

\bibitem{Kazhdan_2020_AIMLAI}
Dmitry Kazhdan, Botty Dimanov, Mateja Jamnik, Pietro Liò, and Adrian Weller.
\newblock {Now You See Me (CME): Concept-based Model Extraction}.
\newblock In {\em Proceedings of the AIMLAI Workshop at the ACM International
  Conference on Information and Knowledge Management}, 2020.

\bibitem{kim2018interpretability}
Been Kim, Martin Wattenberg, Justin Gilmer, Carrie Cai, James Wexler, Fernanda
  Viegas, et~al.
\newblock {Interpretability Beyond Feature Attribution: Quantitative Testing
  with Concept Activation Vectors (TCAV)}.
\newblock In {\em Proceedings of the International Conference on Machine
  Learning (ICML)}, pages 2668--2677, 2018.

\bibitem{kingma2014adam}
Diederik~P Kingma and Jimmy Ba.
\newblock {Adam: A Method for Stochastic Optimization}.
\newblock {\em arXiv preprint arXiv:1412.6980}, 2014.

\bibitem{koh_2020_ICML}
Pang~Wei Koh, Thao Nguyen, Yew~Siang Tang, Stephen Mussmann, Emma Pierson, Been
  Kim, and Percy Liang.
\newblock {Concept Bottleneck Models}.
\newblock In {\em Proceedings of the International Conference on Machine
  Learning (ICML)}, pages 5338--5348. PMLR, 2020.

\bibitem{Kumar_CVPR_2009}
Neeraj Kumar, Alexander~C. Berg, Peter~N. Belhumeur, and Shree~K. Nayar.
\newblock {Attribute and Simile Classifiers for Face Verification}.
\newblock In {\em Proceedings of the IEEE International Conference on Computer
  Vision (CVPR)}, pages 365--372, 2009.

\bibitem{lampert2009learning}
Christoph~H Lampert, Hannes Nickisch, and Stefan Harmeling.
\newblock {Learning to Detect Unseen Object Classes by Between-class Attribute
  Transfer}.
\newblock In {\em Proceedings of the IEEE Conference on Computer Vision and
  Pattern Recognition (CVPR)}, pages 951--958, 2009.

\bibitem{lipton2017doctor}
Zachary~C Lipton.
\newblock {The Doctor Just Won't Accept That!}
\newblock {\em arXiv preprint arXiv:1711.08037}, 2017.

\bibitem{liu2020deep}
Yuan Liu, Ayush Jain, Clara Eng, David~H Way, Kang Lee, Peggy Bui, Kimberly
  Kanada, Guilherme de Oliveira~Marinho, Jessica Gallegos, Sara Gabriele,
  et~al.
\newblock A deep learning system for differential diagnosis of skin diseases.
\newblock {\em Nature Medicine}, 26(6):900--908, 2020.

\bibitem{lopez2017skin}
Adria~Romero Lopez, Xavier Giro-i Nieto, Jack Burdick, and Oge Marques.
\newblock {Skin Lesion Classification from Dermoscopic Images using Deep
  Learning Techniques}.
\newblock In {\em Proceedings of the IEEE International Conference on
  Biomedical Engineering (BioMed)}, pages 49--54, 2017.

\bibitem{Lucieri_IJCNN2020}
Adriano Lucieri, Muhammad~Naseer Bajwa, Stephan~Alexander Braun, Muhammad~Imran
  Malik, Andreas Dengel, and Sheraz Ahmed.
\newblock On interpretability of deep learning based skin lesion classifiers
  using concept activation vectors.
\newblock In {\em Proceedings of the International Joint Conference on Neural
  Networks (IJCNN)}, pages 1--10, 2020.

\bibitem{lucieri2022exaid}
Adriano Lucieri, Muhammad~Naseer Bajwa, Stephan~Alexander Braun, Muhammad~Imran
  Malik, Andreas Dengel, and Sheraz Ahmed.
\newblock {ExAID: A Multimodal Explanation Framework for Computer-Aided
  Diagnosis of Skin Lesions}.
\newblock {\em Computer Methods and Programs in Biomedicine}, page 106620,
  2022.

\bibitem{lucieri2020explaining}
Adriano Lucieri, Muhammad~Naseer Bajwa, Andreas Dengel, and Sheraz Ahmed.
\newblock {Explaining AI-based Decision Support Systems using Concept
  Localization Maps}.
\newblock In {\em Proceedings of the International Conference on Neural
  Information Processing (NeurIPS)}, pages 185--193, 2020.

\bibitem{Majkowska_Radiology2020}
Anna Majkowska, Sid Mittal, David~F Steiner, Joshua~J Reicher, Scott~Mayer
  McKinney, Gavin~E Duggan, Krish Eswaran, Po-Hsuan Cameron~Chen, Yun Liu,
  Sreenivasa~Raju Kalidindi, et~al.
\newblock {Chest Radiograph Interpretation with Deep Learning Models:
  Assessment with Radiologist-adjudicated Reference Standards and
  Population-adjusted Evaluation}.
\newblock {\em Radiology}, 294(2):421--431, 2020.

\bibitem{PH2}
Teresa Mendonça, Pedro~M. Ferreira, Jorge~S. Marques, André R.~S. Marcal, and
  Jorge Rozeira.
\newblock {PH2 - A Dermoscopic Image Database for Research and Benchmarking}.
\newblock In {\em Proceedings of the International Conference of the IEEE
  Engineering in Medicine and Biology Society (EMBC)}, pages 5437--5440, 2013.

\bibitem{Nachbar1994JAAD}
Franz Nachbar, Wilhelm Stolz, Tanja Merkle, Armand~B Cognetta, Thomas Vogt,
  Michael Landthaler, Peter Bilek, Otto Braun-Falco, and Gerd Plewig.
\newblock {The ABCD Rule of Dermatoscopy: High Prospective Value in the
  Diagnosis of Doubtful Melanocytic Skin Lesions}.
\newblock {\em Journal of the American Academy of Dermatology}, 30(4):551--559,
  1994.

\bibitem{pennington2014glove}
Jeffrey Pennington, Richard Socher, and Christopher~D Manning.
\newblock {GloVe: Global Vectors for Word Representation}.
\newblock In {\em Proceedings of the Conference on Empirical Methods in Natural
  Language Processing (EMNLP)}, pages 1532--1543, 2014.

\bibitem{Rudin_NMI2019}
Cynthia Rudin.
\newblock {Stop Explaining Black Box Machine Learning Models for High Stakes
  Decisions and Use Interpretable Models Instead}.
\newblock {\em Nature Machine Intelligence}, 1(5):206--215, 2019.

\bibitem{sarkar_2022_CVPR}
Anirban Sarkar, Deepak Vijaykeerthy, Anindya Sarkar, and Vineeth~N
  Balasubramanian.
\newblock {A Framework for Learning Ante-hoc Explainable Models via Concepts}.
\newblock In {\em Proceedings of the IEEE/CVF Conference on Computer Vision and
  Pattern Recognition (CVPR)}, pages 10286--10295, 2022.

\bibitem{HAM10000}
Philipp Tschandl, Cliff Rosendahl, and Harald Kittler.
\newblock {The HAM10000 Dataset: A Large Collection of Multi-Source
  Dermatoscopic Images of Common Pigmented Skin Lesions}.
\newblock {\em Scientific Data}, 5(1):1--9, 2018.

\bibitem{wickramanayake2021comprehensible}
Sandareka Wickramanayake, Wynne Hsu, and Mong~Li Lee.
\newblock {Comprehensible Convolutional Neural Networks via Guided Concept
  Learning}.
\newblock In {\em Proceedings of the IEEE International Joint Conference on
  Neural Networks (IJCNN)}, pages 1--8, 2021.

\bibitem{xiang2019towards}
Alec Xiang and Fei Wang.
\newblock {Towards Interpretable Skin Lesion Classification with Deep Learning
  Models}.
\newblock In {\em Proceedings of the AMIA Annual Symposium}, page 1246, 2019.

\bibitem{Xu_Engineering2021}
Yesheng Xu, Ming Kong, Wenjia Xie, Runping Duan, Zhengqing Fang, Yuxiao Lin,
  Qiang Zhu, Siliang Tang, Fei Wu, and Yu-Feng Yao.
\newblock {Deep Sequential Feature Learning in Clinical Image Classification of
  Infectious Keratitis}.
\newblock {\em Engineering}, 7(7):1002--1010, 2021.

\end{thebibliography}
}

\end{document}